\documentclass[preprint,12pt]{elsarticle}

\usepackage{amssymb}
\usepackage{multirow}
\usepackage{xcolor}

\journal{}

\begin{document}

\begin{frontmatter}



\title{Exploring the Impact of Generative Artificial Intelligence in Education: A Thematic Analysis}


\author[inst1]{Abhishek Kaushik\corref{cor1}\fnref{label2}}
\fntext[label2]{Corrosponding Author}
\affiliation[inst1]{organization={School of Informatics and Creative Arts, Dundalk Institute of Technology},
       addressline={Dublin Road}, 
           city={Dundalk},
         postcode={A91 K584}, 
            state={Co. Louth},
            country={Ireland}}
\author[inst1]{Sargam Yadav}

\author[inst7]{Andrew Browne}
\author[inst3]{David Lillis}
\author[inst7]{David Williams}

\author[inst1]{Jack Mc Donnell}
\author[inst1]{Peadar Grant}
\author[inst1]{Siobhan Connolly Kernan}
\author[inst8]{Shubham Sharma}

\author[inst9]{Mansi Arora}

\affiliation[inst7]{organization={Dublin Business School},
       addressline={13/14 Aungier St.}, 
       city={Dublin},
            postcode={D02 WC04}, 
            state={Co. Dublin},
            country={Ireland}}

\affiliation[inst3]{organization={University College Dublin},
         addressline={Belfield}, 
         city={Dublin},
          postcode={D04C1P1}, 
      state={Co. Dublin},
           country={Ireland}}

\affiliation[inst8]{organization={Technological University Dublin},
         addressline={Dublin 7 },
         city={Dublin},
          postcode={D07 H6K8}, 
      state={Co. Dublin},
           country={Ireland}}

\affiliation[inst9]{organization={Jagan Institute of Management Studies},
           addressline={3, Institutional Area, Sector-5, Rohini}, 
            city={Delhi},
            postcode={110085}, 
            state={Delhi},
            country={India}}
\begin{abstract}
The recent advancements in Generative Artificial intelligence (GenAI) technology have been transformative for the field of education. Large Language Models (LLMs) such as ChatGPT and Bard can be leveraged to automate boilerplate tasks, create content for personalised teaching, and handle repetitive tasks to allow more time for creative thinking. However, it is important to develop guidelines, policies, and assessment methods in the education sector to ensure the responsible integration of these tools. In this article, thematic analysis has been performed on seven essays obtained from professionals in the education sector to understand the advantages and pitfalls of using GenAI models such as ChatGPT and Bard in education. Exploratory Data Analysis (EDA) has been performed on the essays to extract further insights  from the text. The study found several themes which highlight benefits and drawbacks of GenAI tools, as well as suggestions to overcome these limitations and ensure that students are using these tools in a responsible and ethical manner. 
\end{abstract}



\end{frontmatter}


\section{Introduction}
\label{sec:intro}

The accelerated advancements in Artificial Intelligence (AI) over the past decade have disrupted several fields such as education \cite{holmes2019artificial}, healthcare \cite{cascella2023evaluating} \cite{yadav2021simplify}, finance \cite{li2023large}, and law \cite{cui2023chatlaw}. Natural Language Processing (NLP) is a subfield of AI responsible for understanding, synthesizing, and generating human language \cite{bird2009natural}. Examples of applications of NLP include sentiment analysis in various languages \cite{kaur2019cooking} \cite{shah2020opinion}, hate speech detection \cite{yadav2023hate} \cite{yadav2024leveraging}, machine translation \cite{stahlberg2020neural}, and question answering \cite{allam2012question}.NLP systems have evolved from early rule-based chatbots, such as ALICE \cite{wallace2009anatomy} and ELIZA \cite{weizenbaum1966eliza}, to the advanced transformer-based systems \cite{vaswani2017attention} such as Bidirectional Encoder Representations from Transformers (BERT) \cite{devlin2018bert} and A Robustly Optimized BERT Pretraining Approach (RoBERTa) \cite{liu2019roberta}, which have found numerous applications \cite{yadav2023comparative} \cite{yadav2022contextualized}.These models are pre-trained on large amounts of data and consist of parameters in the millions or billions, enabling them to capture the context of the conversation and other linguistic complexities \cite{radford2019language} \cite{devlin2018bert}. They have gained popularity due to their ability to enhance human productivity, boost creativity \cite{lee2022rethinking}, and support personalized and continuous learning \cite{doe2022personalized}.

Generative AI (GenAI) refers to AI systems capable of creating text, audio, and images, in response to user prompts \cite{glotzbach2024generative}. In recent years, the outstanding capabilities of GenAI tools and LLMs such as ChatGPT \cite{chatgpt}, Bing-AI \cite{bing_ai}, and Bard \cite{bard} have highlighted the potential of this technology in education \cite{katz2024gpt}. The ability of 
ChatGPT to carry out natural sounding conversations and respond in the style requested by the user can be harnessed to develop engaging teaching aids that suit the needs of the students \cite{khosrawi2022conversational} \cite{grassini2023shaping}. For software development, students can use AI-assisted coding tools to generate boilerplate templates \cite{becker2023programming}, perform troubleshooting and debugging \cite{mayer2002can}, and generate documentation \cite{wang2022documentation}. GenAI tools can function as a personalized tutor for students, encouraging an adaptive learning environment, and reducing their dependence on educators \cite{holmes2019artificial} \cite{luckin2016intelligence}. OpenAI's website has provided a student's guide to writing with ChatGPT, which suggests use cases such as formatting citations, providing foundational knowledge on a new topic, providing relevant research sources, providing answers to specific questions, providing tailored and iterative feedback, and suggesting counterarguments for a thesis \cite{guide}. However, there are several pitfalls associated with GenAI technology that can lead to concerns about academic integrity, plagiarism \cite{tian2023gptzero} \cite{pudasaini2024survey}, over-reliance on and potential misuse of the technology, and transparency about their operation \cite{zawacki2019systematic}. Therefore, it is important to understand and address these challenges.


Thematic analysis is a qualitative research approach used to identify themes and patterns from data \cite{alhojailan2012thematic} \cite{peel2020beginner}. It involves generating initial codes from the data, aggregating similar codes together, and drawing insights from the resulting themes. Exploratory Data Analysis (EDA) is a data analytics process that also aims to uncover patterns in relationships in a dataset.


In this study, opinions, in the form of unstructured essays, were obtained from 7 educators discussing the potential benefits and challenges of integrating GenAI in education. Thematic analysis has been performed on these essays by extracting codes and deriving themes from them. Additionally, EDA has been performed on the text to derive insights from the essays. The rest of the paper is structured as follows: Section \ref{sec:motivation} details the motivation of the study along with the hypothesis and research questions. Section \ref{sec:methodology} covers the methodology used to conduct the study and perform thematic analysis on educator opinions. Section \ref{sec:educatoropinions} includes the opinion essays provided by the 7 educators. In Section \ref{sec:thematicanalysis}, the identified themes are discussed in detail, and Section \ref{sec:eda} details the results of EDA. Section \ref{sec:discussion} attempts to answer the research questions in context of the findings of the analysis. Section \ref{sec:conclusion} concludes the study.

\section{Motivation}
\label{sec:motivation}
The impressive capabilities of GenAI tools such as their ability to carry out natural conversations about a wide array of topics \cite{brown2020language}, perform analysis on multimodal data \cite{chatgpt}, and generate personalized content \cite{liu2021deep}, come with several risks. Although these tools can be greatly beneficial by serving functions such as automating repetitive tasks \cite{baker2016educational}, and providing personal tutoring \cite{law2024application}, they pose significant ethical concerns and can be detrimental to the learning process and development of problem-solving skills \cite{resnick2024generative}. The motivation behind conducting this study is to gain insight from educator opinions about the use of GenAI in the field of education. The individual perspectives of the educators can be a helpful tool in understanding the potential advantages and challenges of this transformative technology. This can help in the effective and ethical integration of GenAI tools in educational practices to harness their potential, while avoiding potential misuse and the limitations presented by the technology.

\subsection{Hypothesis and Research Questions}
The hypothesis for this study is as follows:

\textit{Hypothesis:} Educators perceive both advantages and challenges in the integration of GenAI in education.

The research questions formulated to explore the hypothesis are as follows:

\begin{enumerate}
    \item RQ1: What potential advantages of GenAI in education are uncovered through thematic analysis of educator opinions?
    \item RQ2: What potential limitations and challenges of GenAI in education are uncovered through thematic analysis of educator opinions?
    \item RQ3: What are the findings of exploratory data analysis on opinion essays?
\end{enumerate}

\section{Methodology}
\label{sec:methodology}

In this section, the methodology used to perform thematic analysis and exploratory data analysis has been discussed.

\begin{table}[!h]
    \centering
    \begin{tabular}{|cc|}
    \hline
    \textbf{No.} & \textbf{Gender} \\
    \hline
       1  &  M \\
        2 & M \\
        3 & F \\
        4 & M \\
        5  & M \\
        6 & F \\
        7 & M \\
       \hline
    \end{tabular}
    \caption{Educator Details}
    \label{tab:educator_details}
\end{table}

Table \ref{tab:educator_details} lists the gender details of the educators who participated in the study. 5 out of the 7 educators are male, which is a high gender imbalance. Educators have provided lectures in machine learning, digital marketing, programming, databases, distributed systems, statistics, game development, and research methods in machine learning. 

\subsection{Thematic Analysis}
\begin{figure}[!h]
    \centering
    \includegraphics[width=0.9\linewidth]{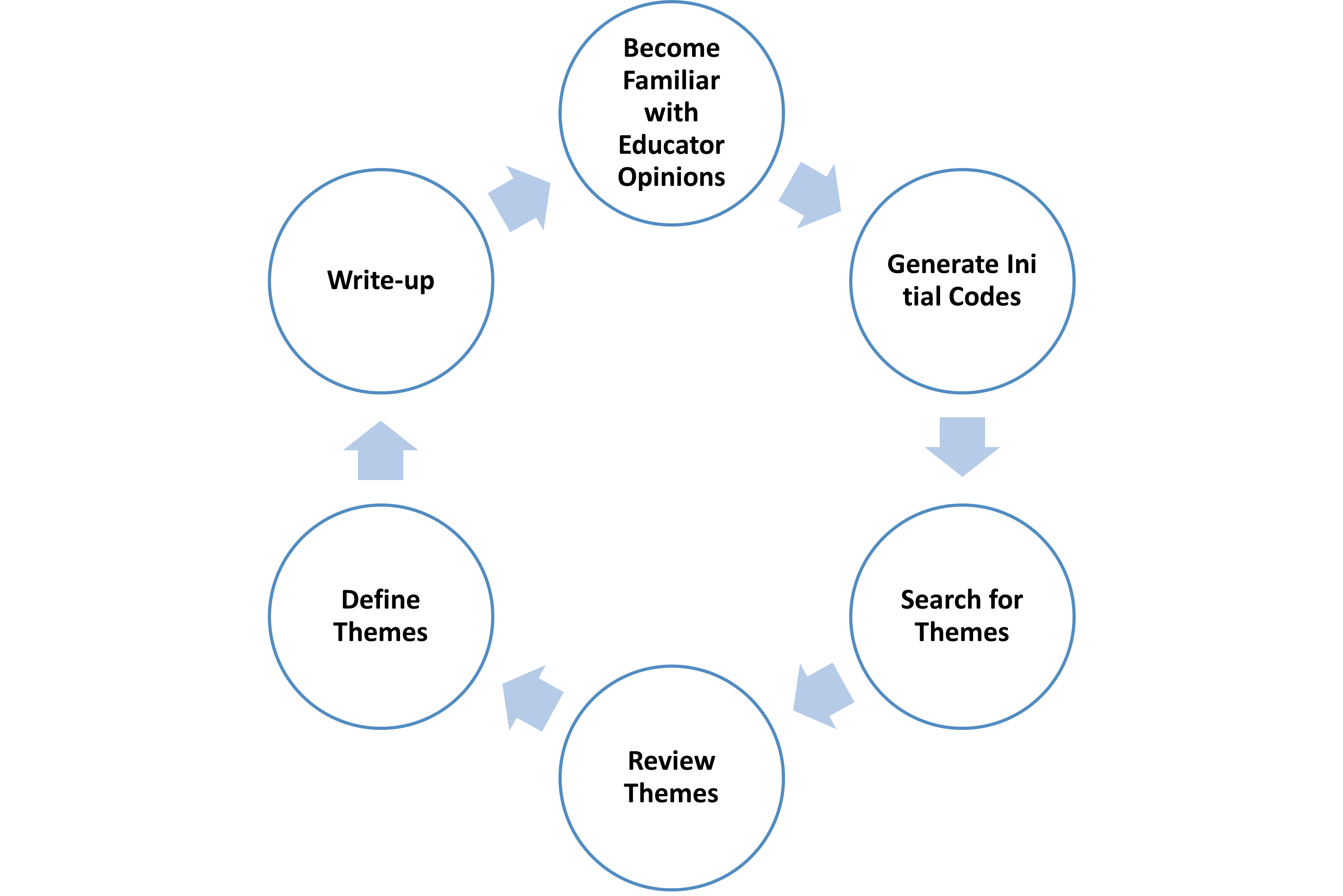}
    \caption{Thematic Methodology}
    \label{fig:them_meth}
\end{figure}

Thematic analysis is a technique to find patterns and themes within qualitative data to uncover underlying topics and ideas  \cite{maguire2017doing}. Figure \ref{fig:them_meth} displays the steps involved in performing thematic analysis as detailed by Braun and Clarke \cite{braun2006using}. It consists of the following steps: Familiarize yourself with the data, generate initial codes from the data, search for themes, review themes, define themes, and complete the write-up. 

\subsection{Exploratory Data Analysis}
In order to perform EDA on the opinion essays, pre-processing has been performed by converting the text to lowercase, removing all stopwords, and lemmatizing the tokens. All the references, images, and headings were removed from the essays. The most common words and bigrams and extracted from the text. 


 \section{Educator Opinions}
\label{sec:educatoropinions}
In the following subsections, the opinions essays provided by the educators have been  included.
\subsection{Educator 1}

Students should understand that AI chatbots are tools/resources that can help them but cannot do everything. For example, an MSc student was interested in studying a topic but did not have an appropriate dataset. They asked ChatGPT to simulate data for them. The dataset that ChatGPT created was nonsensical and highly inappropriate to answer the questions they wanted to examine. AI chatbots are not able to create simulated data without clear and explicit instructions. A possible exercise in data management and study design would be to ask AI chatbots to simulate data. Writing specific instructions to create data with an appropriate structure could be a useful exercise. Asking chatbots the right questions is the skill that needs to be learned.
Concerns related to plagiarism and academic misconduct are valid in my opinion. Even though third level institutions are putting policies in place to deter students from claiming the work of AI chatbots is their own, use is still prevalent, and often difficult to detect or to prove. Several of my colleagues teaching mathematics at third level have noted that many students do not have the patience to learn mathematics. They are used to instant answers from online calculators and AI chatbots. The art of taking time to figure out a problem has been lost. This is worrying as one of the main attributes of maths graduates is problem-solving skills. Universities also need to work with primary and secondary schools so that students are not dependent on AI chatbots when they start third level education. 

It is important that students are taught about possible biases in AI-generated content. In many cases, the methodology for producing content is not transparent or easily accessed and the relevance or accuracy of the information must be questioned. It has also been shown that there are concerns around copyright issues when using AI chatbots \cite{lucchi2024chatgpt}. Students need to be educated about the potential dangers of this. A module or course on AI chatbots could be a mandatory part of every third-level degree as part of core skills to make sure students are informed about the use of such tools.

I am teaching programming to a group studying for a master’s degree in data analytics. The lectures are lab-based with a focus on solving practical problems in class. Almost all the students immediately open AI chatbots to help them with the exercises. This can help with minor fixes, but when it is used to write full functions it removes the learning to independently solve problems. It is often the case that the chatbot has written code close to correct, but students do not question the output, and without developing the skills to write functions themselves they are unable to correct and improve the chatbot generated code. Possible exercises and assignments could involve taking human or AI generated code that is partially correct and adapting it.

Universities and students should be careful about adopting the use of AI-driven tools. Students could be frustrated if lecturers use chatbots, but they are not allowed to. 
It is important to educate students on the weaknesses of ChatGPT. For example, it regularly miscalculates simple arithmetic operations.
AI chatbots are excellent at relaying facts and writing text but reduce the possibility for creativity from the learner. There is an inherent struggle when writing an essay or code that I think is a necessary struggle. It is necessary to learn techniques and problem-solving skills and it is necessary to write creatively and to grapple with new concepts.
People say it is like when the calculator was introduced – it will become normalised and an accepted part of education. However, as someone with over ten years of experience as an educator at third level, I have seen a very poor standard of mental maths and an over-reliance on calculators. Students could do with having better arithmetic skills in my opinion. Students with better abilities of estimation are better equipped to identify when an answer is clearly wrong and not blindly accept the calculator’s answer. Similarly, an over-reliance on AI chatbots will reduce students’ ability to write clearly and think independently. They will be less able to critique the output from AI chatbots which is by no means perfect.

\subsection{Educator 2}
Artificial Intelligence was consigned by many to either an academic or science fiction curiosity. Although the founding of the MIT AI lab predates the internet's inception, Artificial Intelligence has remained largely a niche research pursuit even inside academia. This changed in 2023 when Generative AI and particularly Open AI's Generative Pre-trained Transformer (GPT) Large Language Models (LLMs) attracted significant interest, popularity and familiarity amongst the general public. The conversational interface of ChatGPT introduced many to constructing prompts and refining output for the first time. It was clear that students were ahead in uptake of ChatGPT in particular than their educators!  
The archetypical computer science education centres on programming. Assistance was largely confined to initial code template generation and basic refactoring tools, mainly centred around languages such as Java and C\# that structurally suited them. Just as Google is supplemented by domain-specific search tools, ChatGPT's generality has now been augmented by tools such as GitHub CoPilot for programming. These are now integrated into modern code authoring tools, and even traditional text editors such as emacs have interface packages available.
As well as programming languages, the computing ecosystem houses a multitude of text-based information: configuration files, Infrastructure-as-code and system administration scripts. I have found that students in diverse fields such as cloud computing, data storage technologies and data architecture have been able to leverage generative AI to produce boilerplate templates. More usefully they can generate minimal working examples from which to develop and integrate their own solutions, reducing the barrier to entry of many tools, and increasing the breadth of their skillset.
Early internet search engines included many operators to fine-tune searches, and whilst Google still supports them, very few users actively take advantage of them. The usefulness of output from GPT models is highly correlated to the quality of the prompts given. Learners will benefit significantly if prompt construction is integrated into information search and retrieval tutorials at an early stage. More specifically, computing students need to see appropriate use of generative AI in coding contexts by their instructors, just as they would encounter the use of refactoring tools by example. Optimal ways to use revolutionary new tools, and knowing when not to use them, is best achieved by experiential practice, not avoidance!
Educators are grappling with the impact that generative AI has had on assessment, particularly highlighted by academic integrity concerns. Many essay-type assessments are at risk of being largely the work of LLMs rather than the student, including perhaps some assessments that were not fit-for-purpose in any event. Practical skill demonstration under examination conditions will probably need to form an increased part of the assessment for many applied subjects, with prohibition or explicit limits on the use of generative AI and other tooling.

\subsection{Educator 3}

After seeing first hand the impact ChatGPT can have on a student that is struggling with getting code to work in a project, ChatGPT was freely available there to fix any bugs the student was struggling with and allowed them to move on to the next part of the project without having to ask for assistance from a supervisor/lecturer. This is an invaluable resource that allows the student more independence in their learning when it should be independent learning. However this only works well when the student has already achieved the foundational learning needed in the area and is now trying to do apply more advanced techniques. The student then has enough knowledge to understand when the prompts it has given ChatGPT has actually lead to a coherent and correct answer. 

Academic integrity has been an issue since for over a century and will continue to be an issue with the current education and research structures \cite{fishman2016academic}. During the pandemic and since the pandemic,  learning has moved to a more blended online learning environment. Universities were already needing to update the policies and procedures to take into account this more fluid learning environment whilst maintaining the integrity of the grades being achieved without the formal onsite externally invigilated exams. They use of these more freely available AI tools has just accelerated this need even more so than the pandemic whether the learning is primarily in class room or online. 

The need to be more inventive with the forms of assessment are needed that even if a student is to use an AI tool, despite explicated prohibited, that you can assess whether the learning has been achieved. This might take place in many different ways whether it be; Q\&A sessions with the students on a topic/project, screencasts where the student explains the work, students have to critique the work of AI tools, etc. But it does mean what has worked in the past to assess this module may not still work now and needs a lot of thought from individual lecturers and programme teams to understand what will work for their courses, ideally guided by updated institutional academic integrity policies.  Formal onsite exams still have a place in this new age of learning, and it might seem like an easy solution to assessing the learning from a student without the use of AI tools. Although for many courses, in particular ICT sector of education, formal onsite exams have long been replaced with various continuous assessment strategies and formal onsite exams should not be brought back in light of these AI challenges after it was argued that is not appropriate way to measure the student’s learning in the area previously. The need for a diverse set of assessment strategies are the best way to assess the students abilities \cite{yu2023reflection}.

Difficult thing to do as pandemic has hindered the learning for a lot of students so given the timeframe currently I think this needs to be looked at but when the pandemic can be isolated out of the data. 

\subsection{Educator 4}

The adoption of any new automation technology is fraught with potential
for pitfalls, misunderstandings and misapplications. Before turning
attention to Large Language Models (LLMs) such as ChatGPT, I would
first choose a more straightforward illustrative example.

\subsubsection{Originality Detection}
Even the least tech-savvy of students and educators have some grasp of how
originality detectors such as Turnitin\footnote{https://turnitin.com/} operate. On a high level, the system
has access to a vast database of text samples (both those gathered online
and those submitted to the system in the past) and newly-submitted work
is compared against this. Text passages that match items in the database
are identified and a ``similarity score” is output.
Even in this relatively straightforward scenario, misinterpretation and
misapplication abounds. Firstly, tools of this type are often deceptively
described as ``plagiarism detection” \cite{meo2019turnitin}, leading to an over-reliance on a
single tool as an arbiter of what constitutes plagiarism.
As noted by Meo and Talha \cite{meo2019turnitin}, plagiarism comes in many forms and
plagiarism detection is an academic judgment. ``Word-for-word plagiarism”
(which originality checkers can effectively discover) is only one aspect. Students
who are compelled to submit their work through such systems often
come to conflate ``similarity score” with ``plagiarism score”. Particularly
in situations where students can see these scores and resubmit their work,
a perception can grow that rephrasing the offending matching sections is
sufficient to avoid plagiarism. Where reworded ideas have been taken from
other sources, without attribution, a heavily-plagiarised document can yield
a low similarity score.
Conversely, a relatively high similarity score does not necessarily constitute
plagiarism either, and it is incumbent on educators to bear this in mind.
There are myriad reasons why sections may match text from a database,
particularly quotations and bibliographies. A submitted work that is overreliant
on lengthy quotations without commentary may be of low quality,
but if cited correctly it does not constitute plagiarism. Originality checkers
should only be used as a tool to identify potential cases of a specific form of
plagiarism, with a human investigation necessary to verify whether or not
this is the case.

In summary, even an understandable tool of this type can directly contribute
to students misunderstanding the concepts of plagiarism, and overzealous
educators making accusations of academic misconduct based on a misinterpretation
of the significance of the evidence to hand.

\subsubsection{Large Language Models}
The role and capabilities of LLMs such as ChatGPT and Bard are much
more difficult to understand, and as such the challenges of dealing with them
in an educational setting are even more pronounced.
Firstly, we should endeavour to understand, even on a basic level, how
a LLM operates. In essence, it learns patterns and relationships between
words, sentences and paragraphs in text, having been exposed to enormous
quantities of human-created text to learn from. Then, given a “prompt”
from a user, it generates text in response, beginning by matching the context
of the prompt against its text store.
As it generates the text, it uses a probabilistic approach to choose words
one at a time. Based on the text it has generated thus far, it tries to predict
what the next word should be. However, to avoid generating the same text
in response to the same prompt each time, an element of randomness is
introduced so as not to always choose the most likely word. Finally, it has a
stopping mechanism that will cause the generation to end as appropriate \cite{wolfram2023chatgpt}.
One other aspect is that ChatGPT is also trained on actual chat logs
between humans, and so it exhibits elements of personality. It is polite to a
fault, apologises for perceived mistakes and appears eager to please.
This leads to another observation, relating to the language that people
use to describe their characteristics, and indeed their shortcomings. Because
of the human-like nature of the generated text, people seem to be
happy to attribute human-like explanations. It has been widely observed
that ChatGPT will generate plausible-looking, incorrect references when
asked to provide a bibliography \cite{alkaissi2023artificial}. Other types of referencing errors have
also been observed (e.g. in law \cite{dixon2023my}). Such errors are typically described as
``hallucinations”, giving them a distinctly human characteristic that implies
real intelligence. Contrast this with a hypothetical AI image classifier that,
presented with a photograph of a cat, predicts that it is a spaceship. In the
latter situation, users are more likely to dismiss the tool’s effectiveness as
being simply wrong.

The human-like characteristic of ChatGPT ultimately means that users
are more likely to trust that its output is correct.
An additional issue is that ChatGPT is innumerate. Although it can
recognise where a numeric value would be appropriate in the text, the specific
value often bears no resemblance to the correct answer. When challenged,
it will attempt to ``correct” the answer (even for relatively straightforward
calculations) and offer an alternative. It is notable that when GPT-
4 recognises that a numeric value is required (at present, the free version
of ChatGPT is based on the earlier GPT-3.5), it will generate a Python
program to perform the calculations, which is a significant advancement.
Students may be attracted by reports of ChatGPT passing the bar exam,
for example \cite{katz2024gpt}, and be tempted to employ it to cheat on university assignments.
Due to the limitations outlined above, strategies such as requiring
correct referencing, or in some cases complex calculations, may result in indications
that the work is not that of the student. A careless student who
simply copy/pastes a ChatGPT-generated essay may find that they have
submitted substandard work, even if their use of LLMs cannot be proven.

\subsubsection{Detection of LLM-Generated Content}

Educators are understandably concerned at the rise in the use of ChatGPT
among students to write essays and assignments. This has led to the launch
of a number of products that claim to be able to differentiate AI-generated
content from human-generated text. Examples include GPTZero\footnote{htps://gptzero.me} \cite{tian2023gptzero} and
ZeroGPT\footnote{https://zerogpt.com}.
To be fair to the creators of these products, their websites are open about
the role their tools are intended to play, and give some detail about how they
are created. For example, GPTZero’s website states the following:
`We test our models on a never-before-seen set of human and
AI articles from a section of our large-scale dataset, in addition to a smaller set of challenging articles that are outside its training distribution.'
ZeroGPT’s website states the following:
`Finally, we employ a comprehensive deep learning methodology, trained on extensive text collections from the internet, educational datasets, and our proprietary synthetic AI datasets produced using various language models.'
Both therefore claim strong accuracy in differentiating between text that
is 100\% AI-generated and text that is 100\% human-generated.
However, as with the originality detection software discussed above, it
is imperative that educators understand what these tools are designed to
do and what they are not designed to do. Only the laziest of students will
directly submit a 100\% AI-generated piece of work. These tools have not
been trained on any dataset that includes proactive efforts to fool them.
In some cases, even the addition of a single space can cause a ChatGPT
detection tool to be fooled \cite{cai2023evade}. Similarly, since AI-generated text does not
contain spelling or grammatical errors, some trivial manipulations can cause
the detection software to fail.
This serves to emphasise some inherent challenges in dealing with the
problem of students using LLMs to complete their assignments. Certainly,
no LLM detector should be relied upon as definitive evidence of wrongdoing,
nor can it definitively exonerate a suspected student. It remains an
open question as to whether a reliable AI-detection tool is even possible. At
best, an educator may use these in a similar way to originality checkers: a
first pass to find suspicious cases that may merit further investigation. However,
human judgment and old-fashioned mechanisms like oral examinations
should remain part of the process.

\subsection{Educator 5}
ChatGPT is a specific software application built on top of Generative AI technology, particularly large language models (LLMs). Generative AI is a broad term that refers to any type of artificial intelligence that can create new content. This can include text, images, music, code, and more. Among these, Large Language Models (LLMs) stand out as a specialized subset of Generative AI, specifically engineered for text generation. LLMs represent a class of artificial intelligence proficient in both text generation and comprehension. They undergo extensive training on extensive datasets containing text and code, enabling them to grasp the intricacies of human language patterns. LLMs find application across a diverse range of tasks, including \cite{lee2022coauthor} :
\begin{itemize}
    \item Text generation: LLMs can generate text, such as news articles, poems, code, scripts, musical pieces, email, letters, etc.
\item Translation: LLMs can translate languages from one language to another.
\item Question answering: LLMs can answer questions in a comprehensive and informative way, even if they are open ended, challenging, or strange.
\item	Summarization: LLMs can summarize long pieces of text into shorter, more concise pieces.
\item Code generation: LLMs can generate code in a variety of programming languages.
\end{itemize}

This has given rise to Ethical and Privacy Concerns around AI generated content in education. Central to understanding the impact past the hype phase is promoting a broader understanding of what these models really are, and how they are designed. Bard\footnote{https://bard.google.com/}, ChatGPT, and Bing\footnote{https://copilot.microsoft.com/chats/} AI are all examples of publicly available large language models (LLMs) that can generate text, translate languages, write different kinds of creative content, and answer your questions in an informative way. While large language models open up many possibilities, there is still much to learn about how people will interact with them \cite{yuan2022wordcraft}. While Bard,  Bing  and ChatGPT all aim to give human-like answers to questions, each performs differently. Bing starts with the same GPT-4 tech as ChatGPT but goes beyond text and can generate images. Bard uses Google's own model, called LaMDA, often giving less text-heavy responses \cite{rahaman2023ai} \cite{crawford2023artificial}. Bard is trained on a dataset of text and code that is specifically designed to improve its dialogue and coding abilities. ChatGPT is trained on a dataset of text that is more general in nature. This means that Bard is better at understanding and responding to natural language, while ChatGPT is better at generating creative text formats. 

Consideration must be given to alignment with educational values. This should ensure AI tools align with educational goals and values, such as critical thinking, creativity, and ethical decision-making. The presence of AI-generated content presents new and unique ethical considerations. Firstly, the very notion of authorship blurs, as AI lacks the capacity for genuine creative ownership. Assigning sole credit to authors who merely provide prompts for AI outputs is equally disingenuous. Therefore, establishing transparent attribution guidelines becomes essential. Secondly, the specter of bias is evident throughout AI research, as AI algorithms can unwittingly mirror societal prejudices present in their training data. Mitigating this necessitates employing diverse datasets and vigilantly monitoring outputs for discriminatory content. Thirdly, the potential for manipulating or fabricating information through AI-generated content, exemplified by deepfakes, poses a significant threat. Safeguards emphasizing factual accuracy and transparency are essential to combat this. Finally, the emotional impact of AI content cannot be ignored. Educators must carefully consider the potential psychological effects, particularly on vulnerable populations within educational settings. In conclusion, navigating the ethical minefield surrounding AI-generated content requires a multifaceted approach, encompassing clear attribution, diverse training data, robust safeguards against misinformation, and thoughtful consideration of the emotional impact on users. By addressing these ethical and privacy concerns, we can ensure AI-generated content and chatbots contribute positively to the educational experience, fostering a safe, responsible, and enriching learning environment.

\begin{figure}
    \centering
    \includegraphics[width=0.8\linewidth]{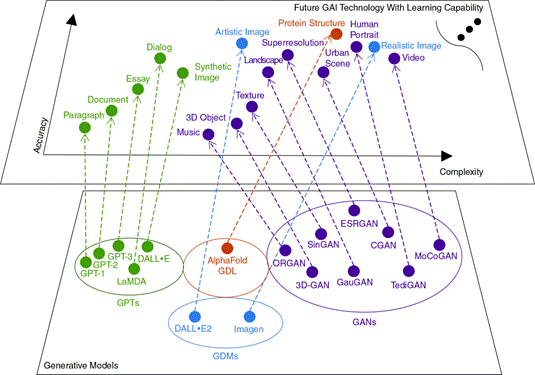}
    \caption{The GAI landscape: generative models and artifacts \cite{jovanovic2022generative}}
    \label{fig:gai}
\end{figure}

\subsection{Educator 6}

Artificial Intelligence (AI), has been increasingly used these days as the prime disposition to take decision, solve problems, write reports and so on and so forth claiming to replace human intelligence in future. Since, it is typically performance based, executing the commands generously without any perceptions or misconceptions of its abilities, it is being increasingly demanded. But another facet to pointed is that, it is only a function of the human command programmed to function with a set pattern or methods and will deliver similar results, sometimes overlapping with the same methodical approach. The future of Artificial Intelligence in such light seems crippled without human intelligence. The future of decision-making, problem-solving lies with a careful concoction of number crunching, big data analysis, research tools and interdisciplinary research which requires correct amount of Human intelligence at every stage, leading to formation of Augmented Intelligence. It is an amalgamation of reckoning the correct ingredients or combination with intuitive abilities of human judgement when equipped with the methodical skill set of Artificial intelligence \cite{dede2021intelligence}.
It is also known as Intelligence Augmentation (IA) or Cognitive augmentation is a new age marriage between man and machine.
AI and IA together have a plan to pen down the future differently when used with a collaborative approach. While AI has been increasingly posing a threat to replace humans, but when it comes to the judgment and the reckoning aspect, we see human stepping in for an informed disposition of intelligence. Instead of avoiding or making attempts to accept this inevitable change, it is now time to look at the aspect as a JV between Humans and Computer. The IA approach shall bring together advances, modernisation and speed in the work approach across business enterprise, Institution, organisation, students, workers and media communities. The idea is to make the most of it by equipping and training the human intelligence with its correct and appropriate use. The tasks which are tedious for humans or repetitive in nature and has redundant value can be done by the AI bots, thus removing the human errors and biases. While the tasks which require interpretations, visionary approach holistic mind set and decree can be done by humans with larger efficiency due to save time at hand. 

The statistics have shown that IA leads to 99\% accuracy in decision making, leading to enhanced productivity. Alexa or other similar bots help you to take commands, recognize voices and eliminate the trouble of remembering and in some cases doing of mundane tasks.
Students have although been using it as convenient tool to plagiarize their creativity lowering the scope of thinking. The University experts have now started to incorporate the Chatbot as the assignment providers to the students, where students are the ones evaluating the assignments created by bots \cite{ICRC2023}. The method is a clever precautionary approach rather than being a cure to the plagiarism.
Unlike straight automation, IA shall enhance cognitive abilities. It’s the IA technology who has to evolve with the open human mind set creating and consuming content with the help of AI, leaving no room for error and creating a powerful and strengthened approach. It uses the strengths of both man and machine while mitigating the risks and threats. The data has repeatedly shown that in organisation, where AI was kept as the sole leader, the human participation was ultimately asked in 30\% of cases \cite{TelusInternational}. In the times of the uncertainty of the VUCA world, this only seems to rise.
If your process has continuous human input, the change management in terms of adopting AI as a peer or colleague to work along will become smooth function of any organisation, While the students community need to be evaluators to understand deprivation from the AI approaches, so that they use it like an equipment rather an a subordinate.
The IA can redefine the landscape of Human performance with harmonious function of partnership between man and machine building a realm of AI powered humans, who increase effectiveness at workplace by opening new horizons of ideas, backed by rationale and vision.

\subsection{Educator 7}
Assuring the veracity of student outputs has always been of concern, but recent developments in Generative AI (Gen AI) have thrown a curve ball at the processes already in place. Lecturers and administrators across our college have been challenged with the double concern of how to embed Gen AI into our teaching as a tool but also assure that it is not misused in producing outputs at the assessment level. The usage of these tools by students was at first met with apprehension but then excitement as it was seen as another important tool within the modern student's arsenal; it has become increasingly apparent that these tools will be and are being used across industry \cite{GenAI2024} so we have noticed that we would be seriously disadvantaging students by not including their usage in the teaching programmes. Early trials are in place for using Gen AI as a part of module teaching and assessment in some modules. But we also need to put in place mechanisms to help prevent their misuse in outputs at assessment level. The Exams Office along with Programme Coordinators have developed changes to overall assessment that reflect the need to be aware of the misuse of AI through more authentic forms of assessment. This discussion deals with some of the early ideas and mechanisms proposed and developed around both of these issues as we both battle and welcome AI.

Many scholars are exploring the ethical usage of Gen AI in the classroom. Some of this research finds high intention by students to use these tools \cite{jain2024gen}. But the perceived usefulness of these tools is questioned in university settings contrasting other research in the area \cite{chan2023ai} which the authors say may be due to a lack of understanding of these tools. Within our own organization we are going to extreme attempts to show students and staff the usefulness of these tools in the classroom as well as their coursework. One such method has involved staff CPD to instill effective usage as well as guidance from our corporate headquarters (Kaplan Inc.). We have also been developing guidance at the Quality Assurance Level for staff and students while some staff members are actively including Gen AI tools in their teaching.. 

Another interesting study looks at Gen AI adoption across the generations. \cite{chan2023ai} found that Generation Z students showed an interest in using Gen AI as a tool in their educational pursuits while Generation X and Y teachers showed optimism towards the tool while expressing concerns about its application. One could indeed map the rise of Gen AI on many other tools that over the years would have seemed to be cheapening the learning experience (the computer to the page, the page to rhetoric…). We have found this mixture of interest and apprehension across our own staff as we learn to deal with this interesting new tool. 

The elephant in the room though, is plagiarism or academic impropriety (the other AI). Having experience working adjunct in two other universities has allowed this researcher multiple viewpoints into the issue as it has arisen. Working in different departments (Humanities and Social Sciences) has also highlighted interesting and varied approaches. The first observation was that Humanities departments, heavily reliant on the the essay form such as Literature Studies, initially showed an aggressive zero-tolerance approach to its usage in coursework while the Social Sciences such as Communications Studies, showed a more balanced approach, recognising it as a tool but still leaning strongly towards penalizing students for its usage as opposed to actively incorporating its usage. It was in our Business College where we found a more balanced approach and this may be due to the prevalence of project-type work which facilitates the ethical usage of Gen AI but also makes its misuse less easy to apply. The nature of project work is a more authentic type of assessment that involves more interaction between the lecturer and student, which makes Gen AI content more obvious in final productions.

The answer to the negative aspects of Gen AI that leads to academic impropriety is to embrace more authentic forms of assessment like the above. To move away from the essay from and towards more regulated and monitored project-type work that also encourages the use of Gen AI as a tool in that process. At the HECA Research Conference 2023, Gen AI and authentic assessment was the centre of many interesting discussions. Indeed, the conference ended with Danielle Logan Fleming of Griffith University, Australia: A message of HOPE: Generative AI and Authentic Interactive Oral Assessment\footnote{https://heca.ie/heca-research-conference-2023-sharing-an-open-research-landscape/}. The idea generated here is that we need much more authentic assessment in our programmes and that we can easily battle the misuse of Gen AI by creating more authentic assessment that engages in a conversation with the students as they develop their work \cite{ward2023interactive}. In short, Gen AI is the future of industry and education and we need to embrace it in our classrooms and our assessment. It cannot be ignored, nor should it, as any institution that attempts to ride out the storm of Gen AI will fall behind and drag their students with them.

\section{Thematic Analysis}
\label{sec:thematicanalysis}

In this section, the results of the thematic analysis approach by Braun and Clarke \cite{braun2006using} have been discussed. The framework helped in the identification of 11 themes, which have been discussed in the following subsections. The themes identified were: `Academic Integrity and Challenges in Assessment',  `Limitations and Misuse of Generative AI'  `The Importance of Prompt Construction', `Critical Thinking and Problem-Solving Skills', `Bias, Transparency, and Ethical Concerns', `Responsible use of GenAI', `GenAI for Programming', `Technical Details of AI Tools', `Advantages of GenAI', `Challenges of GenAI', and `Miscellaneous'.

\begin{figure}[!h]
    \centering
    \includegraphics[width=1\linewidth]{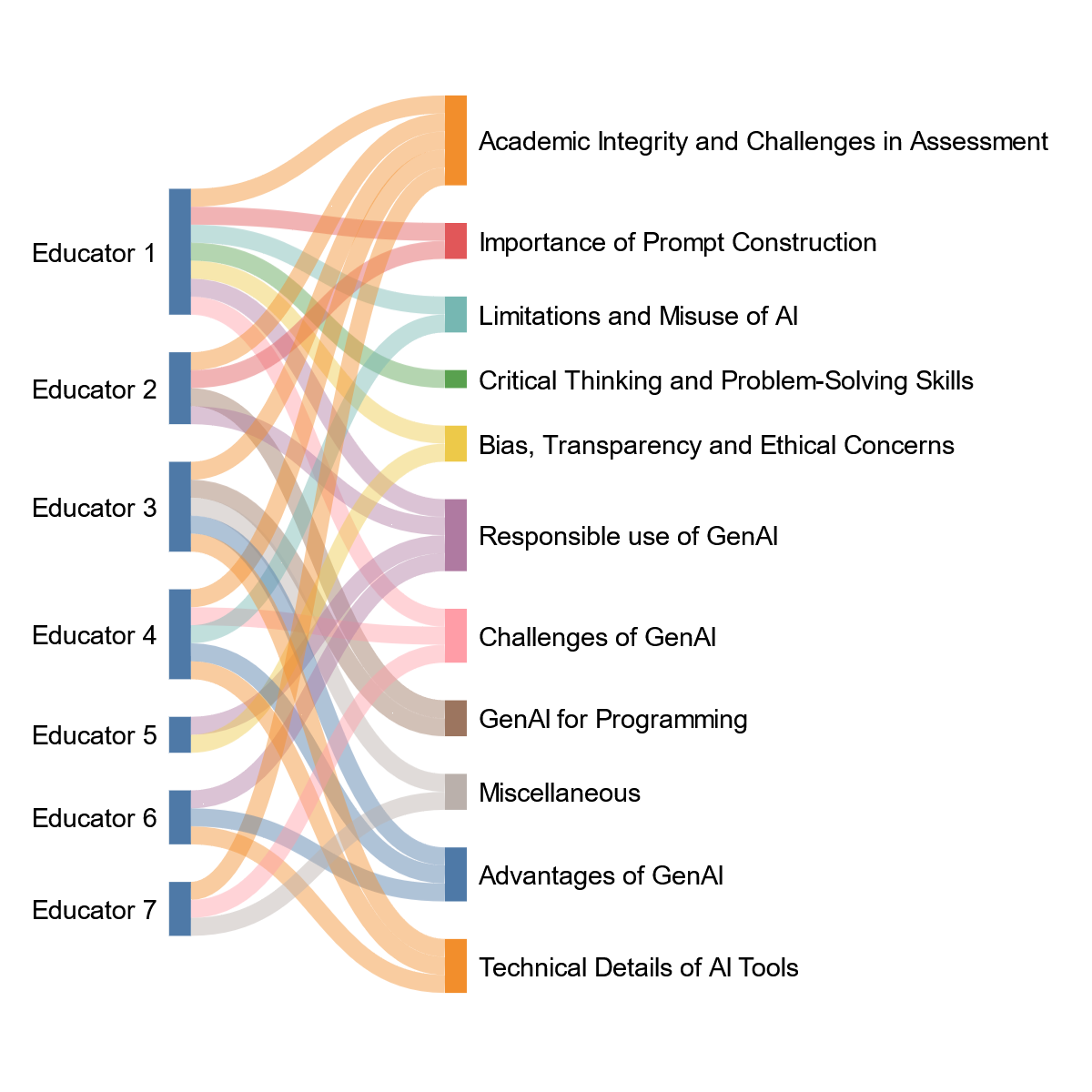}
    \caption{Thematic mapping}
    \label{fig:them_mapping}
\end{figure}

Figure \ref{fig:them_mapping} displays a sankey diagram that illustrates the themes discussed by each educator. `Academic Integrity and Challenges in Assessment' is the most prevalent theme, indicating that it is a significant concern amongst educators. Other prevalent themes include `Responsible use of GenAI' discussed by 4 educators, and `Challenges of GenAI', `Technical Details of AI Tools' and `Advantages of GenAI', each discussed by 3 educators. Themes such as `Importance of Prompt Construction' and `Bias, Transparency, and Ethical Concerns' have been specifically discussed by a minority of the educators. Certain educators, such as educators 1 and 2, have discussed a variety of themes.


\subsection{Academic Integrity and Challenges in Assessment}
`Academic Integrity and Challenges in Assessment' is the most commonly discussed theme, mentioned by 5 out of the 7 educators.

Table \ref{tab:codes_academic_integrity} lists the final codes for the theme `Academic Integrity and Challenges in Assessment' for each educator. Plagiarism can become rampant due to the free availability of GenAI tools, and AI-generated content can be difficult to detect \cite{jawahar2019does}. Developing tools to detect AI generated content can be challenging \cite{tian2023gptzero}, if possible at all, and some can be evaded by simply adding a single space \cite{cai2023evade}. The development of new and innovative assessment methods, such interactive oral assessment \cite{ward2023interactive}, project-based work, and peer evaluations \cite{sesay2024redefining}, is essential.

\begin{table}[!ht]
    \centering
    \begin{tabular}{|c|p{12cm}|}
    \hline
       \textbf{Edc.}  & \textbf{Codes}  \\
         \hline
         1 &  Concerns about plagiarism and academic misconduct, AI generated work difficult to detect\\
         2 & Challenges with AI’s impact on assessments and academic integrity, essay type assessments at risk, importance of practical skill demonstration \\
         3 &Academic Integrity is an issue, academic policies must be updated to ensure academic integrity, freely available AI tools further impact integrity, Inventive and diverse assessment methods needed, alternative assessment methods importance of updating institutional academic integrity policies, \\
         6 &Misuse of AI by students for plagiarism, students evaluate AI-generated assignments to work-around plagiarism \\
         7 &Generative AI worsened the issue of plagiarism detection, assessment methods must be updated to prevent misuse, increased interaction between the lecturer and student for assessment, monitored project-type work encourages ethical use of Gen A \\
         \hline
    \end{tabular}
    \caption{Codes for Academic Integrity and Challenges in Assessment}
    \label{tab:codes_academic_integrity}
\end{table}

\subsection{Limitations and Misuse of Generative AI}
The theme `Limitations and Misuse of Generative AI' includes codes that discuss general limitations and the potential misuses of these tools, which could not be aggregated into a single theme. Table \ref{tab:codes_limitaions} lists the final codes for the theme for the educators who discussed it. Educator 4 discusses how ChatGPT hallucinates plausible sounding references \cite{alkaissi2023artificial} and creates other referencing errors \cite{dixon2023my}. ChatGPT also struggles with basic arithmetic \cite{bubeck2023sparks}. 

\begin{table}[!ht]
    \centering
    \begin{tabular}{|c|p{12cm}|}
    \hline
        \textbf{Edc.}  &  \textbf{Codes}  \\
         \hline
         1 &Heavy reliance by students on AI chatbots for coding, Limitations of AI chatbots as tools, need to educate students about limitations of ChatGPT, ChatGPT often miscalculates simple arithmetic operations  \\
         3 & Foundational knowledge necessary for effective use of ChatGPT\\
         4 &ChatGPT generates incorrect references, ChatGPT is innumerate \\
         \hline
    \end{tabular}
    \caption{Codes for Limitations and Misuse of AI}
    \label{tab:codes_limitaions}
\end{table}

\subsection{The Importance of Prompt Construction}
The theme `Importance of Prompt Construction' consists of codes that discuss the importance of constructing high-quality and precise prompts to obtain relevant and accurate responses from a GenAI tool. Table \ref{tab:codes_Prompt} lists the final codes for the theme for the educators who discussed it. Educators 1 and 2 suggest supplying GenAI chatbots with precise prompts will provide optimal results \cite{liu2023pre}, and tutorials for prompt construction should be included in the curriculum \cite{holmes2019artificial}. 

\begin{table}[!ht]
    \centering
    \begin{tabular}{|c|p{12cm}|}
    \hline
        \textbf{Edc.}  &  \textbf{Codes}  \\
         \hline
         1 & Explicit instructions required to generate simulated data, precise and correct prompts must be constructed; \\
         2 &  Introduction to prompt engineering through ChatGPT, Importance of high-quality prompts for GPT, need for including information search and retrieval tutorials\\
         \hline
    \end{tabular}
    \caption{Codes for The Importance of Prompt Construction}
    \label{tab:codes_Prompt}
\end{table}

\subsection{Critical Thinking and Problem-Solving Skills}

The theme `Critical Thinking and Problem-Solving Skills' consists of codes which discuss the impact of GenAI technology on the critical thinking skills of students. Table \ref{tab:codes_critical} lists the final codes for the theme for the educators who discussed it. GenAI tools can both promote and hinder the development of these skills \cite{resnick2024generative}. Educator 1 mentions particular concerns such as decline in creativity and independent thinking due to over-reliance on GenAI tools \cite{george2023preparing}.

\begin{table}[!ht]
    \centering
    \begin{tabular}{|c|p{12cm}|}
    \hline
        \textbf{Edc.}  &  \textbf{Codes}  \\
         \hline
         1 & Decline in problem-solving skills due to AI; AI chatbots diminish problem-solving skills in coding; AI chatbots reduce student creativity, struggle essential for problem solving skills, over-reliance on AI chatbots will reduce independent thinking;  \\
           \hline
    \end{tabular}
    \caption{Codes for Critical Thinking and Problem-Solving Skills}
    \label{tab:codes_critical}
\end{table}

\subsection{Bias, Transparency, and Ethical Concerns}
Table \ref{tab:codes_bias} lists the final codes for the theme `Bias, Transparency, and Ethical Concerns' for the educators who discussed it. The theme `Bias, Transparency, and Ethical Concerns' consists of codes that touch on biases in AI -generated content \cite{farrelly2023generative}, copyright concerns with AI generated content \cite{lucchi2023chatgpt}, and authorship debates around such content \cite{crawford2023artificial}.
The lack of transparency in the development and deployment of these tools is also a significant barrier in their integration in the educational curriculum \cite{zawacki2019systematic}. Educator 5 has also expressed concerns regarding the proliferation of AI-generated fake content, which can have a detrimental impact on misinformation in politics and journalism \cite{vaccari2020deepfakes} \cite{chesney2019deepfakes}. The potential psychological impact that AI generated content can have on the students, which can be positive or negative, must also be kept in mind \cite{luckin2016intelligence}.

\begin{table}[!ht]
    \centering
    \begin{tabular}{|c|p{12cm}|}
    \hline
        \textbf{Edc.}  &  \textbf{Codes}  \\
         \hline
         1 &Students need to learn about biases in AI-generated content, lack of transparency in generative AI methodologies, copyright concerns with AI generated content, \\
         
         5 & Ethical and Privacy Concerns around AI generated content in education, ethical considerations such as authorship, bias, and AI generated fake content, need for safeguards for accuracy and transparency, potential psychological impact of AI generated content must be considered \\
         \hline
    \end{tabular}
    \caption{Codes for Bias, Transparency, and Ethical Concerns}
    \label{tab:codes_bias}
\end{table}

\subsection{Responsible use of GenAI}
In the theme `Responsible use of GenAI', suggestions for ethical and responsible use of GenAI have been highlighted. Table \ref{tab:codes_responsible} lists the final codes for the theme for the educators who discussed it.
Educator 1 suggests possible changes that can be made to class assignments that are given to students. Educators 1 and 2 suggest that the proper use of GenAI tools must be taught as part of the curriculum. Educator 3 highlights the importance of studying user interaction with LLMs \cite{yuan2022wordcraft}. Educator 6 highlights the importance of human oversight in utilizing GenAI.

\begin{table}[!ht]
    \centering
    \begin{tabular}{|c|p{12cm}|}
    \hline
        \textbf{Edc.}  &  \textbf{Codes}  \\
         \hline
         1 &A course on AI chatbots in education, Possible assignments can adapt partial correct human or AI generated code;  \\
         2 & Appropriate use of generative AI should be instructed; \\
          3 &User interaction with LLMs has to be studied\\
          6 & AI ineffective without human intelligence, creative problem solving can be performed by humans,  continuous human input necessary for AI adoption,  \\
          \hline
    \end{tabular}
    \caption{Codes for Responsible use of GenAI}
    \label{tab:codes_responsible}
\end{table}

\subsection{GenAI for Programming}

\begin{table}[!ht]
    \centering
    \begin{tabular}{|c|p{12cm}|}
    \hline
        \textbf{Edc.}  &  \textbf{Codes}  \\
         \hline
         2 & Focus of computer science education on programming, earlier coding assistants limited to template generation and basic tools, current AI tools for programming integrated into text editors; Presence of diverse text-based information in computing environments,generative AI for boilerplate templates and minimal working examples\\
         3 & ChatGPT as a troubleshooting tool for students\\
         \hline
    \end{tabular}
    \caption{Codes for GenAI for Programming}
    \label{tab:codes_programming}
\end{table}

The theme `GenAI for Programming' includes mentions of the benfits of GenAI tools in software development. Table \ref{tab:codes_programming} lists the final codes for the theme for the educators who discussed it. Educator 2 discusses GenAI tools in the context of computer programming, and the tasks, such as template generation and pair programming \cite{ma2023ai}, that can be handled efficiently by these tools. Educator 3 mention the benefits of GenAI as a troubleshooting tool in programming \cite{arora2024analyzing}. Despites the various advantages, the use of AI coding assistants has also led to a concerning decrease in code reuse \cite{harding2024coding}.

\subsection{Technical Details of AI Tools}

The theme `Technical Details of AI Tools' consists of codes that describe the mechanism behind AI tools. Table \ref{tab:codes_technical} lists the final codes for the theme `Technical Details of AI Tools' for the educators who discussed it. Educator 4 describes the mechanism behind originality detectors, and suggests that these tools are completely ineffective in detecting AI-generated content. Educators 4, 5, and 6 also describe the mechanism behind LLMs and ChatGPT, and compare a few examples. 

\begin{table}[!ht]
    \centering
    \begin{tabular}{|c|p{12cm}|}
    \hline
        \textbf{Edc.}  &  \textbf{Codes}  \\
         \hline
         4 & Originality detectors match text passages and generate similarity score, LLMs generate text via a probabilistic approach, ChatGPT trained on actual chat logs\\
         5 &Large Language Models (LLMs) subset of Generative AI, LLMs proficient in text generation and comprehension, LLMs undergo extensive training on large datasets, applications of LLMs, Bing can also generate images, LaMDA gives less text-heavy responses, Bard better at NLU and NLG, ChatGPT better at generating creative text formats \\
         6 & AI programmed by humans\\
         \hline
    \end{tabular}
    \caption{Codes for Technical Details of AI Tools}
    \label{tab:codes_technical}
\end{table}


\subsection{Advantages of GenAI}

The theme `Advantages of GenAI' is composed of codes that detailed advantages of using GenAI in education that were not covered under a single theme. Table \ref{tab:codes_advantages} lists the final codes for the theme for the educators who discussed it. Educator 3 explores the potential of GenAI tools such as ChatGPT in promoting personalized and adaptive learning environments, limiting the immediate need for educators to complex-learning environments \cite{holmes2019artificial} \cite{luckin2016intelligence}. 
Educator 4 discusses how the human-like characteristics of GenAI tools such as politeness inspire trust and credibility amongst the users \cite{lee2004trust}. Intelligence Augmentation (IA) is a concept that focuses on enhancing human capabilities through the use of technology \cite{engelbart2023augmenting}, and has been discussed in detail by Educator 6. 

\begin{table}[!ht]
    \centering
    \begin{tabular}{|c|p{12cm}|}
    \hline
        \textbf{Edc.}  &  \textbf{Codes}  \\
         \hline
         3 &ChatGPT promoting independent learning and reducing reliance on educators \\
         4 & ChatGPT exhibits polite personality, GPT-4 generates Python script when a numeric value is required\\
         6 & Augmented Intelligence future of decision-making and problem-solving, Intelligence Augmentation combination of human intuition and methodical skill set of AI, IA will improve modernisation and speed, repetitive tasks can be performed by AI, AI can reduce human errors and bias\\
         \hline
    \end{tabular}
    \caption{Codes for Advantages of GenAI}
    \label{tab:codes_advantages}
\end{table}

\subsection{Challenges of GenAI}
The theme `Challenges of GenAI' consists of all the concerns related to the integration of GenAI into education that could not be aggregated into a single theme. Table \ref{tab:codes_challenges} lists the final codes for the theme for the educators who discussed it. Educator 1 discusses challenges such as the need to verify AI generated content and code \cite{marcus2019rebooting} and potential frustration amongst students due to inconsistent policies for using GenAI tools \cite{zawacki2019systematic}. Human oversight is essential to ensure the accuracy and reliability of AI-generated content, especially in high-stake situations \cite{Heaven2019WhyDA}. 
Educators 4 and 7 discuss other challenges such as limited perceived usefulness due to lack of knowledge and the skepticism towards new technologies \cite{selwyn2019should}. 

\begin{table}[!ht]
    \centering
    \begin{tabular}{|c|p{12cm}|}
    \hline
        \textbf{Edc.}  &  \textbf{Codes}  \\
         \hline
         1 & Lack of verification of AI-generated code, Caution in adopting AI-driven tools, frustration amongst students due to inconsistent AI use policies,accuracy and relevance of AI-generated content must be verified\\
         4 &challenges in adoption of new technologies, human-like characteristics (hallucinations) attributed to ChatGPT, ChatGPT trusted due to human-like characteristics\\
         7 & concerns about effective integration and prevention of misuse, lack of understanding of GenAI tools can affect perceived usefulness,\\
         \hline
    \end{tabular}
    \caption{Codes for Challenges of GenAI}
    \label{tab:codes_challenges}
\end{table}

\subsection{Miscellaneous}

The `Miscellaneous' theme consists of code that could not be aggregated into other themes. Table \ref{tab:codes_misc} lists the final codes for the theme for the educators who discussed it. Educator 3 discusses the drastic shift towards online learning that occurred during the pandemic \cite{adedoyin2023covid}, and the challenges that were faced in reliably and correctly assessing online tests and assignments \cite{marinoni2020impact}. Educator 7 discusses the variability in adoption of GenAI across different generations and departments of an institute. Age can be a determining factor in an educator's willingness to adopt a GenAI tool \cite{chan2023ai}. While younger and middle-aged educators approach the new tool with optimism and concern, older educators are more skeptical of it, and preferred to maintain control of the key aspects of teaching \cite{seldon2018fourth}. The apprehension demonstrated by educators could be due to a lack of familiarity with the technology, concerns about loss of employment, and resistance to change \cite{selwyn2019should}.

\begin{table}[!ht]
    \centering
    \begin{tabular}{|c|p{12cm}|}
    \hline
        \textbf{Edc.}  &  \textbf{Codes}  \\
         \hline
        3 & Move to a blended learning environment in the pandemic\\ 
        7 & Generation Z students showed interest in Gen AI as a tool, Generation X and Y teachers showed optimism and concern, different departments have varied approaches to GenAI, Humanities departments heavily relied on essay generation, Business College had a more balanced approach due to project-type work \\
         \hline
    \end{tabular}
    \caption{Codes for Miscellaneous}
    \label{tab:codes_misc}
\end{table}

 \section{Exploratory Data Analysis}
In the section, the findings of performing EDA on the opinion essays have been discussed.
 
\label{sec:eda}
\begin{table}[!ht]
    \centering
    \small
    \begin{tabular}{|c|p{10cm}|} 
    \hline
    \textbf{Educator} & \textbf{Terms (Frequency)} \\ \hline
    1 & student (16), chatbots (14), ai (13), data (6), skill (6), could (5), answer (4), exercise (4), third (4), level (4) \\ \hline
    2 & ai (7), tool (7), many (6), generative (6), use (5), student (4), search (4), assessment (4), largely (3), language (3) \\ \hline
    3 & student (13), learning (12), work (6), pandemic (5), ai (5), need (5), formal (4), onsite (4), exam (4), tool (4) \\ \hline
    4 & text (17), student (11), tool (10), plagiarism (10), even (9), chatgpt (9), work (7), may (7), llm (6), originality (6) \\ \hline
    5 & text (12), ai (11), language (10), llm (10), content (9), chatgpt (6), model (6), code (6), ethical (5), bard (5) \\ \hline
    6 & human (17), intelligence (10), ai (9), approach (8), ia (7), student (5), future (5), function (4), set (4), artificial (3) \\ \hline
    7 & ai (23), gen (18), tool (15), student (12), assessment (12), usage (8), authentic (7), also (6), staff (5), across (4) \\ \hline
    \end{tabular}
    \caption{Top 10 most frequent words with count for each educator}
    \label{tab:word_freq}
\end{table}

Table \ref{tab:word_freq} displays the top 10 most frequent words for each educator with their respective counts. Words such as `student', `ai', and `tool' appear in the top 10 most frequent words for almost all educators, indicating that several educators discussed the role of students in the adoption of GenAI tools in education. `pandemic' is amongst the most frequent words for Educator 3, as they discussed the shift towards online learning that occurred during the pandemic. Educator 5 has frequently mentioned examples of LLMs such as `chatgpt' and `bard'. `ia', or intelligence augmentation, has been mentioned 7 times by Educator 6, as they have highlighted several advantages of using GenAI tools in collaboration with human creativity and intelligence. Terms related to the theme `Academic Integrity and Challenges in Assessment', such as `assessment', `plagiarism', and `exam', are frequent across all educators. 

\begin{table}[!ht]
    \centering
    \small
    \begin{tabular}{|c|c|}
    \hline
     \textbf{Educator} &  \textbf{Bigram count} \\
    \hline
      1& ai chatbots (12), third level (4), simulate data (2), possible exercise (2)\\
         \hline
        2 & generative ai (5), artificial intelligence (2), refactoring tool (2), use generative (2) \\
         \hline
       
         3 & formal onsite (4), ai tool (4), onsite exam (3), student struggling (2)\\
         \hline
         4 & similarity score (4), language model (3), originality checker (3), large language (2)\\
         \hline
         5 &large language (4), language model (4), aigenerated content (4), generative ai  (3)\\
         \hline
         6 & human intelligence (4), artificial intelligence (3), man machine (3), ai increasingly (2)\\
         \hline
         7 & gen ai (18), ai authentic (3), ai tool (3), authentic assessment (3)\\
         \hline
    \end{tabular}
    \caption{Top 4 frequent bigrams}
    \label{tab:bigram_frequency}
\end{table}

Table \ref{tab:bigram_frequency} displays the top 4 most frequent bigrams across all educators. Again, bigrams such as `originality checker' and `authentic assessment', which correlate to the theme `Academic Integrity and Challenges in Assessment' are frequent across all educators. 

\begin{table}[!ht]
    \centering
    \begin{tabular}{|ccc|}
    \hline
      \textbf{Educator}  & \textbf{No. of Sentences} & \textbf{Avg. word count per sentence} \\
       \hline
       1 & 39 &  19.59\\
         2 & 20 & 25.15\\
         3 & 15& 36.33\\
         4 & 59 & 24.93\\
         5 & 34 & 21.85 \\
         6 & 27 & 28.52 \\
         7 &  32 & 30.72 \\
         \hline
    \end{tabular}
    \caption{Sentence count and Average word count per sentence}
    \label{tab:avg}
\end{table}

Table \ref{tab:avg} displays the sentence count and average word count per sentence. Educator 4 has the highest number of sentences (59), and the longest opinion essay, indicating that they discussed several themes (4 as can be seen from Figure \ref{fig:them_mapping}). Educator 1 has the second highest number of sentences, indicating that they have made several arguments and discussed several themes (5 as can be seen from Figure \ref{fig:them_mapping}). Educator 3 has the lowest number of sentences (15), but also the highest average word count per sentence (36.33), and has discussed only 2 themes (as can be seen from Figure \ref{fig:them_mapping}). This suggests that the opinions essays vary greatly in length.

\section{Discussion}
\label{sec:discussion}
In this section, the findings of the study will be discussed in context with the research questions and hypothesis. Limitations of the study will also be highlighted, along with future scope of the work. 

\begin{enumerate}
    \item RQ1: What potential advantages of GenAI in education are uncovered through thematic analysis of educator opinions?

The thematic analysis methodology followed in the study has revealed several themes from the opinion essays. Identified themes such as `GenAI for Programming'  and `Advantages of GenAI' indicate both specific and general advantages of GenAI tools in education. A number of AI-powered tools such as GitHub Copilot and Tabnine have been trained on vast amounts of code and can be effectively used to generate boilerplate templates, debug code, and documentation. Within the theme `Advantages of GenAI', other benefits of GenAI tools were discussed such as their ability to provide custom learning environments \cite{holmes2019artificial}, reducing educator's workload  \cite{luckin2016intelligence}, and inherent trustworthiness due to their polite disposition \cite{lee2004trust}. The concept of IA also highlights how GenAI tools can be used to enhance human intelligence and creativity by allowing more time for creative problem solving \cite{engelbart2023augmenting}.

    \item RQ2: What potential limitations and challenges of GenAI in education are uncovered through thematic analysis of educator opinions?

    The thematic analysis methodology also identified several themes which conveyed the various challenges and limitations that can arise when integrating GenAI tools into education. The most commonly discussed themes amongst all educators was `Academic Integrity and Challenges in Assessment', which discussed concerns such as plagiarism, academic misconduct, outdated assessment approaches \cite{ward2023interactive}, etc. The development of tools to detect AI- generated content is challenging \cite{jawahar2019does} \cite{tian2023gptzero}. 
    Another theme that identified a specific challenge was Critical Thinking and Problem-Solving Skills,' where educators discussed the harm that over-reliance on GenAI tools can do to the development of critical thinking skills \cite{resnick2024generative} \cite{george2023preparing}. The theme `Limitations and Misuse of Generative AI' discusses limitations of GenAI technology such as their tendency to hallucinate \cite{alkaissi2023artificial}, the potential for misuse and over-reliance, and the necessity to possess a foundational knowledge of the subject when using these tools. The theme `Bias, Transparency, and Ethical Concerns' discusses the concerns that arise from unintended biases in AI \cite{farrelly2023generative}, potential copyright and authorship conflicts \cite{lucchi2023chatgpt} \cite{crawford2023artificial}, the lack of transparency in the mechanism of GenAI tools, and the potential harm from the misinformation caused by deepfakes and other false generated content \cite{vaccari2020deepfakes}. The theme `Challenges of GenAI' discusses the challenges that need to addressed to ensure responsible and ethical use of GenAI tools such as the need for verification of AI-generated content \cite{marcus2019rebooting}, lack of knowledge about the technology \cite{zawacki2019systematic}, and skepticism towards change \cite{selwyn2019should}.

    \item RQ3: What are the findings of exploratory data analysis on the opinion essays?
    EDA on the educator responses has provided certain insights into the their opinions on the use of GenAI tools in education. The responses were varied in length, ranging form 15 to 59 sentences. After preprocessing and stopword removal, the top 10 words and top 4 bigrams for each essay was highlighted. Some of the most frequent tokens, such as `assessment', `plagiarism', and `exam', align with the identified themes, ìn this case `Academic Integrity and Challenges in Assessment'. 
    
    
\end{enumerate}

Therefore, educators perceive both advantages and drawbacks of GenAI in the education. Despite the various challenges that present themselves and the potential of misuse of these tools, effective policy making, guidance for proper use, and updated assessments methods can allow both students and educators to use these tools ethically. The study was limited in the form of feedback taken. The number of educators was only 7, with only two female educators present in the sample. Analysing opinions of a large number of educators from different demographics may allow for a better insight. 

\section{Conclusion}

\label{sec:conclusion}
The advent and free availability of various GenAI tools and LLMs has the potential to significantly impact traditional educational practices. However, it is important to distinguish genuine concerns about the use of this technology from the hype, so that the necessary policies, laws, and frameworks may be developed to ensure it's responsible integration in the educational sector. In this study, thematic analysis has been performed on opinions essays about the use of GenAI in education obtained from 7 educators. Several themes emerged from this analysis, which highlighted both the potential benefits and limitations of these tools. They can serve as a personal tutor, handle repetitive tasks with ease, and provide better engagement due to their ability to generate human-like text. However, several limitations and challenges also become apparent, such as academic integrity, plagiarism, development of new and well-rounded assessment methods, and ethical and copyright concerns. The most frequent tokens obtained by performing EDA on the opinion essays align with the identified themes. The future scope of this study includes obtaining expert feedback through other approaches (questionnaires, interviews, etc.), performing case studies on the subject, and performing sentiment analysis.

 \bibliographystyle{elsarticle-num} 
 \bibliography{references}

\end{document}